\documentclass[10pt,twocolumn,letterpaper]{article}

\usepackage{cvpr}
\usepackage{times}
\usepackage{epsfig}
\usepackage{graphicx}
\usepackage{amsmath}
\usepackage{amssymb}
\usepackage{multirow}
\usepackage{tabularx}
\usepackage{booktabs}

\usepackage[pagebackref=true,breaklinks=true,letterpaper=true,colorlinks,bookmarks=false]{hyperref}

\cvprfinalcopy % *** Uncomment this line for the final submission

\def\cvprPaperID{} % *** Enter the CVPR Paper ID here

\ifcvprfinal\fi
\begin{document}

%%%%%%%%% TITLE
%\title{Dismystrifying the Large-scale Chest X-ray Disease Reporting Challenges using Raw Retrospective Radiological Data: a Critical Perspective}

\title{TieNet: Text-Image Embedding Network for Common Thorax Disease \\Classification and Reporting in Chest X-rays}

\author{Xiaosong Wang\thanks{Both authors contributed equally.} $^1$, Yifan Peng\footnotemark[1] $^2$, Le Lu $^1$, Zhiyong Lu $^2$,Ronald M. Summers $^1$ \\
	$^1$Department of Radiology and Imaging Sciences, Clinical Center,\\
	$^2$ National Center for Biotechnology Information, National Library of Medicine, \\
	National Institutes of Health, Bethesda, MD 20892\\
%	$^3$ Nvidia Corporation \\
	{\tt\small \{xiaosong.wang,yifan.peng,le.lu,luzh,mohammad.bagheri,rms\}@nih.gov}
}

\maketitle
%\thispagestyle{empty}

%%%%%%%%% ABSTRACT
\begin{abstract}
Chest X-rays are one of the most common radiological examinations in daily clinical routines. Reporting thorax diseases using chest X-rays is often an entry-level task for radiologist trainees. Yet, reading a chest X-ray image remains a challenging job for learning-oriented machine intelligence, due to (1) shortage of large-scale machine-learnable medical image datasets, and (2) lack of techniques that can mimic the high-level reasoning of human radiologists that requires years of knowledge accumulation and professional training. In this paper, we show the clinical free-text radiological reports%, that accompany X-ray images in hospital picture and archiving communication systems, 
can be utilized as a priori knowledge for tackling these two key problems. We propose a novel Text-Image Embedding network (TieNet) for extracting the distinctive image and text representations. Multi-level attention models are integrated into an end-to-end trainable CNN-RNN architecture for highlighting the meaningful text words and image regions. We first apply TieNet to classify the chest X-rays by using both image features and text embeddings extracted from associated reports. The proposed auto-annotation framework achieves high accuracy (over 0.9 on average in AUCs) in assigning disease labels for our hand-label evaluation dataset.
%We then demonstrate the minimal amount of `ground truth' labels (NLP mined disease keywords) required for achieving similar performance as previous rule-based method.
Furthermore, we transform the TieNet into a chest X-ray reporting system. It simulates the reporting process and can output disease classification and a preliminary report together%, with X-ray images being the only input
. The classification results are significantly improved (6\% increase on average in AUCs) compared to the state-of-the-art baseline on an unseen and hand-labeled dataset (OpenI).

\end{abstract}\vspace{-3mm}

%%%%%%%%% BODY TEXT
\section{Introduction}
In the last decade, challenging tasks in computer vision have gone through different stages, from sole image classification to multi-category multi-instance classification/detection/segmentation to more complex cognitive tasks that involve understanding and describing the relationships of object instances inside the images or videos.
The rapid and significant performance improvement is partly driven by publicly accessible of the large-scale image and video datasets with quality annotations, \eg{}, ImageNet~\cite{deng2009imagenet}, PASCAL VOC~\cite{everingham2015pascal}, MS COCO~\cite{lin2014microsoft}, and Visual Genome~\cite{krishna2016visual} datasets.
In particular, ImageNet pre-trained deep Convolutional Neural Network (CNN) models~\cite{jia2014caffe,krizhevsky2012imagenet,lin2014network} has become an essential basis (indeed an advantage) for many higher level tasks, \eg{}, Recurrent Neural Network (RNN) based image captioning~\cite{vinyals2015show,karpathy2017deep,plummer2015flickr30k,gan2017semantic}, Visual Question Answering~\cite{wu2016ask,zhu2016visual7w,yu2017multi,nam2017dual}, and instance relationship extraction~\cite{johnson2016densecap,hu2017modeling,dai2017detecting}.

\begin{figure}[]
	%\vspace*{-1em}
	\includegraphics[width=1\columnwidth]{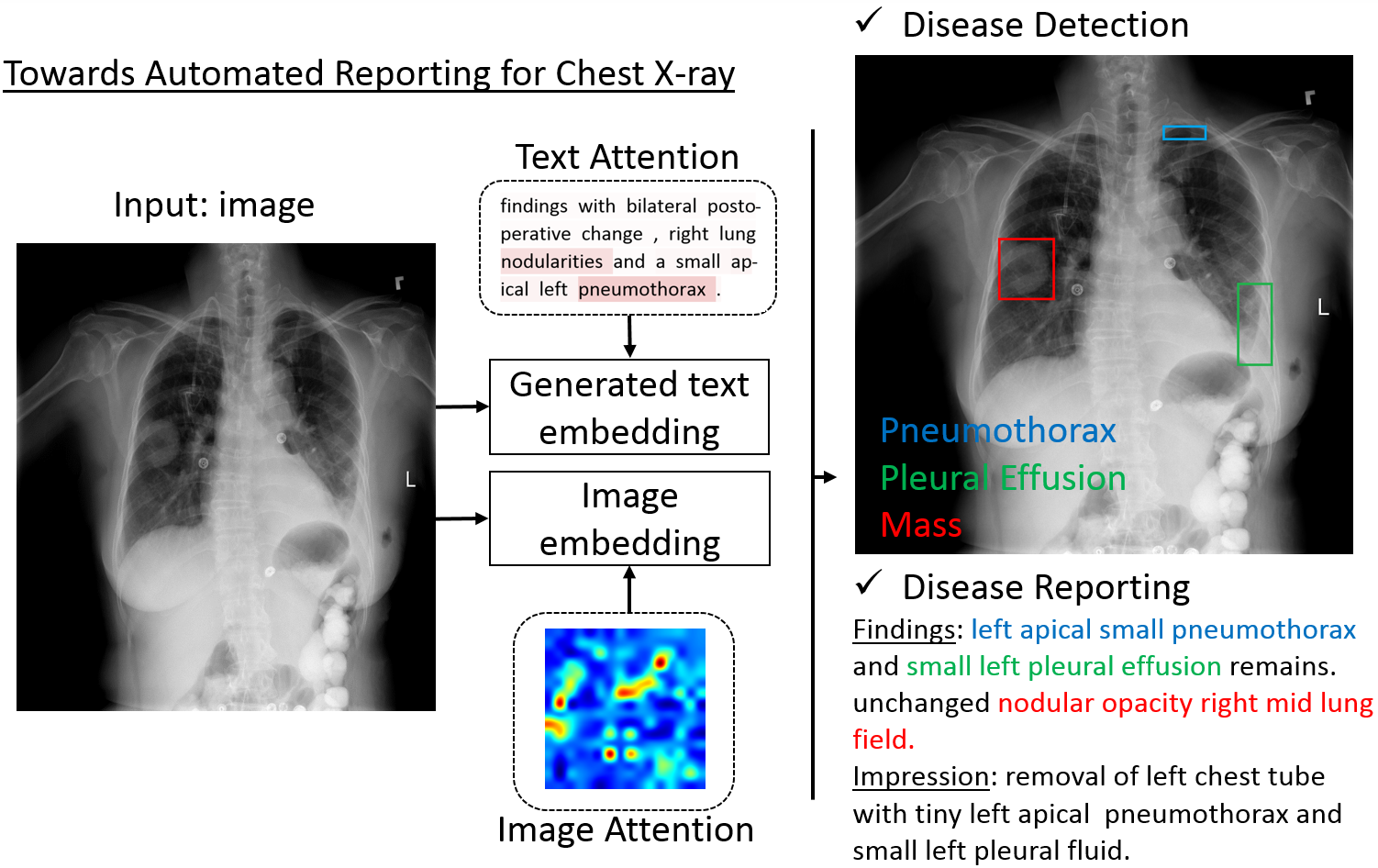}
	\caption{Overview of the proposed automated chest X-ray reporting framework. A multi-level attention model is introduced.% to produce saliency-highlighted text and image embeddings for the input image.
	}
	\label{fig:overview}
	%\vspace*{-1em}
\end{figure}

On the contrary, there are few publicly available large-scale image datasets in the medical image domain.
Conventional means of annotating natural images, e.g
crowd-sourcing, cannot be applied to medical images due to the fact that these tasks often require years of professional training and domain knowledge.
On the other hand, radiological raw data (\eg{}, images, clinical annotations, and radiological reports) have been accumulated in many hospitals' Picture Archiving and Communication Systems (PACS) for decades.
The main challenge is how to transform those retrospective radiological data into a machine-learnable format. Accomplishing this with chest X-rays represents a major milestone in the medical-imaging community~\cite{wang2017chestxa}.

Different from current deep learning models, radiologists routinely observe multiple findings when they read medical images and compile radiological reports.
One main reason is that these findings are often correlated.
For instance, liver metastases can spread to regional lymph nodes or other body parts.
By obtaining and maintaining a holistic picture of relevant clinical findings, a radiologist will be able to make a more accurate diagnosis.
To our best knowledge, developing a universal or multi-purpose CAD framework, which is capable of detecting multiple disease types in a seamless fashion, is still a challenging task.
However, such a framework is a crucial part to build an automatic radiological diagnosis and reporting system.
%
%Thus, this approach differs from the methods that radiologists routinely apply to read medical image studies and compile radiological reports. In latter case, multiple findings are observed and they are often correlated. For instance, liver metastases can spread to regional lymph nodes or other body parts.
%
% However, it remains greatly challenging to develop a universal or multi-purpose CAD framework, capable of detecting multiple disease types in a seamless fashion. Such a framework is crucial to building an automatic radiological diagnosis and reasoning system.  

Toward this end, we investigate how free-text radiological reports can be exploited as \textit{a priori} knowledge using an innovative text-image embedding network. We apply this novel system in two different scenarios.
We first introduce a new framework for auto-annotation of the chest X-rays by using both images features and text embeddings extracted from associated reports.
Multi-level attention models are integrated into an end-to-end trainable CNN-RNN architecture for highlighting the meaningful text words and image regions.
In addition, we convert the proposed annotation framework into a chest X-ray reporting system (as shown in Figure \ref{fig:overview}). The system stimulates the real-world reporting process by outputting disease classification and generating a preliminary report spontaneously.
The text embedding learned from the retrospective reports are integrated into the model as \textit{a priori} knowledge and the joint learning framework boosts the performance in both tasks in comparison to previous state-of-the-art.
%We validate our approach on three different datasets, including the existing large-scale chest X-ray dataset provided by Wang et al.~\cite{wang2017chestxa} through an inter-institutional agreement.
%The resulting classification is significantly improved (3-7\% increase in AUC for every disease class) compared to the state-of-the-art baseline and the Bleu scores for the generated reports are higher than the baseline method.

Our contributions are in fourfold: (1) We proposed the Text-Image Embedding Network, which is a multi-purpose end-to-end trainable multi-task CNN-RNN framework; (2) We show how raw report data, together with paired image, can be utilized to produce meaningful attention-based image and text representations using the proposed TieNet.
%raw reports are not that convenient for publicly sharing due to the difficulty of PHI anonymization while attention-encode sentence embedding is a feasible solution for sharing associated diagnosis info for each image;
(3) We outline how the developed text and image embeddings are able to boost the auto-annotation framework and achieve extremely high accuracy for chest x-ray labeling; %we also demonstrate how many labeled data are required for training such auto-annotation framework; and
(4) Finally, we present a novel image classification framework which takes images as the sole input, but uses the paired text-image representations from training as \textit{a prior} knowledge injection, in order to produce improved classification scores and preliminary report generations.

Importantly, we validate our approach on three different datasets and the TieNet improves the image classification result (6\% increase on average in area under the curve (AUC) for all disease categories) in comparison to the state-of-the-art on an unseen and hand-labeled dataset (OpenI \cite{demner2015preparing}) from other institute. Our multi-task training scheme can help not only the image classification but also the report generation by producing reports with higher BLEU scores than the baseline method.

%some highlights:
%
%attention-encoded sentence embedding
%
%treat attention-encoded sentence embedding as a memory unit, applied to image-based disease classification problem
\begin{figure*}[th]
	\includegraphics[width=\linewidth]{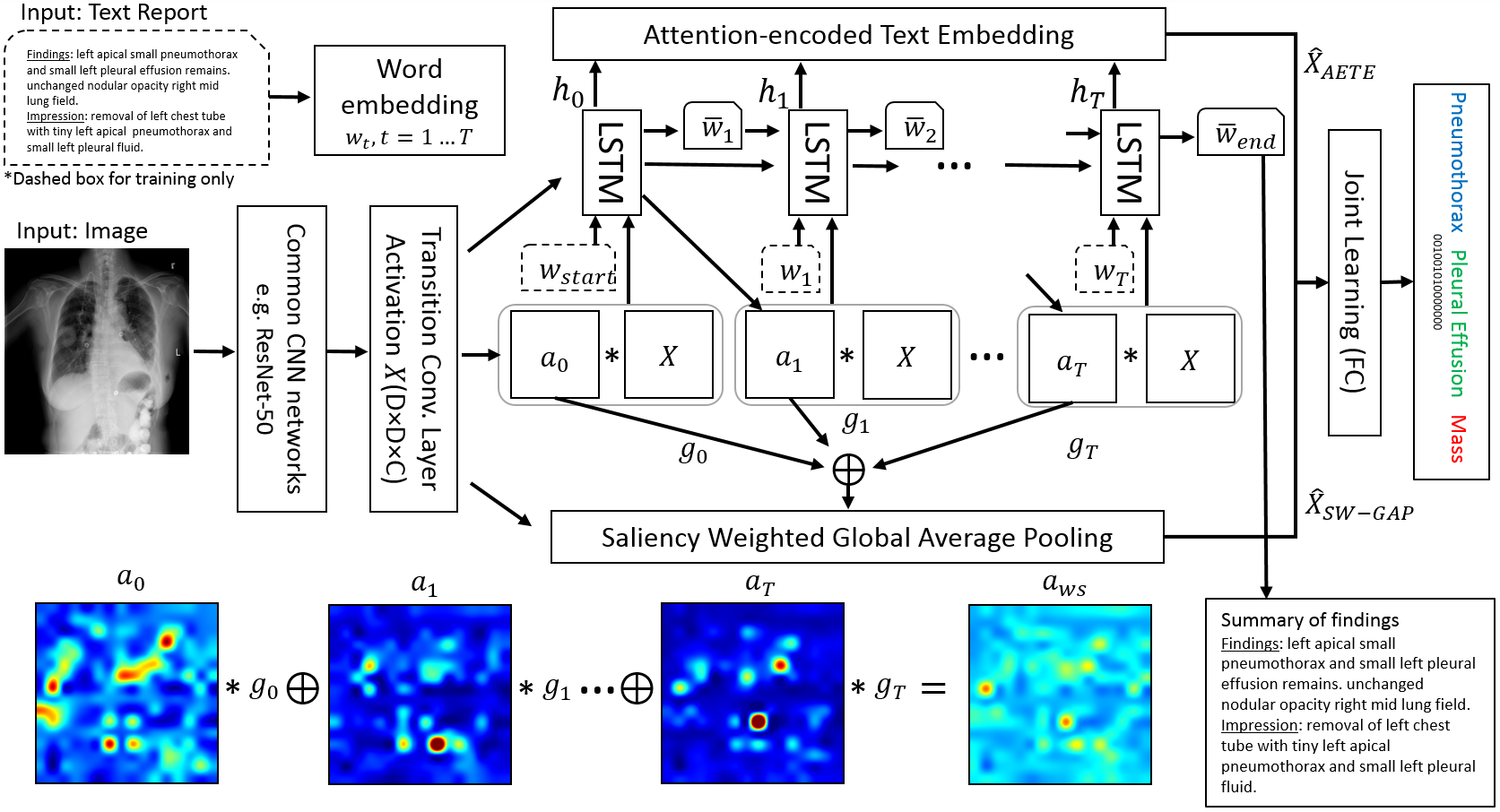}
	\caption{Framework of the proposed chest X-ray auto-annotation and reporting framework. Multi-level attentions are introduced to produce saliency-encoded text and image embeddings.}
	\label{fig:framework}
\end{figure*}
%-------------------------------------------------------------------------
\section{Related work}

Computer-Aided Detection (CADe) and Diagnosis (CADx) has long been a major research focus in medical image processing~\cite{chartrand2017deep}.
In recent years, deep learning models start to outperform conventional statistical learning approaches in various tasks, such as automated classification of skin lesions~\cite{esteva2017dermatologist}, detection of liver lesions~\cite{ben-cohen2016fully}, and detection of pathological-image findings~\cite{zhang2017tandemnet}.
However, current CADe methods typically target one particular type of disease or lesion, such as lung nodules, colon polyps or lymph nodes~\cite{liu2017detection}.

Wang \etal{}~\cite{wang2017chestxa} provide a recent and prominent exception, where they introduced a large scale chest X-ray dataset by processing images and their paired radiological reports (extracted from their institutional PACS database) with natural language processing (NLP) techniques. The publicly available dataset contains $112,120$ front-view chest X-ray images of $30,805$ unique patients \footnote{\url{https://nihcc.app.box.com/v/ChestXray-NIHCC/}}. However, radiological reports contain richer information than simple disease binary labels, \eg{}, disease location and severity, which should be exploited in order to fully leverage existing PACS datasets. Thus, we differ from Wang \etal{}'s approach by leveraging this rich text information in order to produce an enhanced system for chest X-ray CADx.

In vision of visual captioning, our work is closed to \cite{xu2015show,vinyals2015pointer,pedersoli2017areas,yu2017multi,nam2017dual}.
Xu \etal{}~\cite{xu2015show} first introduced the sequence-to-sequence model and spatial attention model into the image captioning task. They conditioned the long short-term memory (LSTM) decoder on different parts of the input image during each decoding step, and the attention signal was determined by the previous hidden state and CNN features.
Vinyals \etal{}~\cite{vinyals2015pointer} cast the syntactical parsing problem as a sequence-to-sequence learning task by linearizing the parsing tree.
Pederoli \etal{}~\cite{pedersoli2017areas} allowed a direct association between caption words and image regions.
More recently, multi-attention models~\cite{yu2017multi,nam2017dual} extract salient regions and words from both image and text and then combine them together for better representations of the pair. 
In medical imaging domain, Shin~\etal{}\cite{shin2016learning} proposed to correlate the entire image or saliency regions with MeSH terms. 
Promising results~\cite{zhang2017mdnet} are also reported in summarizing the findings in pathology images using task-oriented reports in the training. 
%We see our model and theirs as having different granularities. 
The difference between our model and theirs lies in that we employ multi-attention models with a mixture of image and text features in order to provide more salient and meaningful embeddings for the image classification and report generation task.

Apart from visual attention, text-based attention has also been increasingly applied in deep learning for NLP~\cite{bahdanau2015neural,meng2015encoding,rush2015neural}.
It attempts to relieve one potential problem that the traditional encoder-decoder framework faces, which is that the input is long or very information-rich and selective encoding is not possible.
The attention mechanism attempts to ease the above problems by allowing the decoder to refer back to the input sequence~\cite{yulia2015not,lin2017structured,liu2016learning}.
To this end, our work closely follows the one used in~\cite{lin2017structured} where they extracted an interpretable sentence embedding by introducing self-attention. Our model paired both the attention-based image and text representation from training as a prior knowledge injection to produce improved classification scores.

\section{Text-Image Embedding Network}

The radiological report is a summary of all the clinical findings and impressions determined during examination of a radiography study.
A sample report is shown in Figure~\ref{fig:overview}.
It usually contains richer information than just disease keywords, but also may consist of negation and uncertainty statements.
In the `findings' section, a list of normal and abnormal observations will be listed for each part of the body examined in the image.
Attributes of the disease patterns, \eg{}, specific location and severity, will also be noted.
Furthermore, critical diagnosis information is often presented in the `impression' section by considering all findings, patient history, and previous studies.
Suspicious findings may cause recommendations for additional or follow-up imaging studies.
As such, reports consist of a challenging mixture of information and a key for machine learning is extracting useful parts for particular applications.

In addition to mining the disease keywords~\cite{wang2017chestxa} as a summarization of the radiological reports, we want to learn a text embedding to capture the richer information contained in raw reports.
Figure~\ref{fig:framework} illustrates the proposed Text-Image Embedding Network.
We first introduce the foundation of TieNet, which is an end-to-end trainable CNN-RNN architecture. Afterwards we discuss two enhancements we develop and integrate, \ie{}, attention-encoded text embedding (AETE) and saliency weighted global average pooling (SW-GAP). Finally, we outline the joint learning loss function used to optimize the framework.

\subsection{End-to-End Trainable CNN-RNN Model}

%In Section~\ref{sec:baseline}, we describe our baseline encoder-decoder model. We extend this baseline in Section~\ref{sec:attention} with our self-attention mechanism.
As shown in Figure~\ref{fig:framework}, our end-to-end trainable CNN-RNN model takes an image $I$ and a sequence of 1-of-$V$ encoded words.
\begin{equation}
\mathbf{S} = \{\mathbf{w}_1,\ldots,\mathbf{w}_T\}, \mathbf{w}_t \in \mathbb{R}^V \textrm{,}
\end{equation}
where $\mathbf{w}_t$ is a vector standing for a $d_w$ dimensional word embedding for the $t$-th word in the report, $V$ is the size of the vocabulary, and $T$ is the length of the report. The initial CNN component uses layers borrowed from ImageNet pre-trained models for image classification, \eg{}, ResNet-50 (from Conv1 to Res5c).
% * <adam.p.harrison@gmail.com> 2017-11-15T01:33:27.486Z:
% 
% > ResNet-50
% Cite
% 
% ^.
The CNN component additionally includes a convolutional layer (transition layer) to manipulate the spatial grid size and feature dimension. 

Our RNN is based off of Xu \etal{}'s visual image spatial attention model~\cite{xu2015show} for image captioning. The convolutional activations from the transition layer, denoted as $\mathbf{X}$, initialize the RNN's hidden state, $\mathbf{h}_{t}$, where a fully-connected embedding, $\phi(\mathbf{X})$, maps the size $d_{X}$ transition layer activations to the LSTM state space of dimension $d_h$. In addition, $X$ is also used as one of the RNN's input. However, following Xu \etal{}~\cite{xu2015show}, our sequence-to-sequence model includes a deterministic and soft visual spatial attention, $\mathbf{a}_{t}$, that is multiplied element-wise to $\mathbf{X}$ before the latter is inputted to the RNN. At each time step, the RNN also outputs the subsequent attention map, $\mathbf{a}_{t+1}$. 

In addition to the soft-weighted visual features, the RNN also accepts the current word at each time step as input. We adopt standard LSTM units~\cite{hochreiter1997long} for the RNN. The transition to the next hidden state can then be denoted as 
\begin{equation}\label{eq:lstm}
\mathbf{h}_t = LSTM([\mathbf{w}_t,\mathbf{a}_{t},\mathbf{X}],\mathbf{h}_{t-1}) \textrm{.}
% * <adam.p.harrison@gmail.com> 2017-11-15T14:15:43.358Z:
%
% > h_t = LSTM([w_t,a_{t},I],h_{t-1})
%
% ^.
\end{equation}
The LSTM produces the report by generating one word at each time step conditioned on a context vector, \ie{}, the previous hidden state $\mathbf{h}_t$, the previously generated words $\mathbf{w}_t$, and the convolutional features of $\mathbf{X}$ whose dimension is $D\times D\times C$. Here $D=16$ and $C=1024$ denote the spatial and channel dimensions, respectively.
%The feedback of $w_t$ in the state update makes $w_{t+1}$ recursively depend on both $X$ and the entire sequence of words $S_{1:t}=(w_1,\ldots,w_t)$, generated so far.
Once the model is trained, reports for a new image can be generated by sequentially sampling $\mathbf{w}_t \sim p(\mathbf{w}_t|\mathbf{h}_t)$ and updating the state using Equation \ref{eq:lstm}.

The end-to-end trainable CNN-RNN model provides a powerful means to process both text and images. However, our goal is also to obtain an interpretable global text and visual embedding for the purposes of classification. For this reason, we introduce two key enhancements in the form of the AETE and SW-GAP. 

\subsection{Attention Encoded Text Embedding}

To compute a global text representation, we use an approach that closely follows the one used in~\cite{lin2017structured}. More specifically, we use attention to combine the most salient portions of the RNN hidden states. Let $\mathbf{H} = (\mathbf{h}_1,\ldots,\mathbf{h}_T)$ be the $d_{h}\times T$ matrix of all the hidden states. The attention mechanism outputs a $r \times T$ matrix of weights $\mathbf{G}$ as
\begin{equation}
\mathbf{G} = softmax (\mathbf{W}_{s2} \, tanh (\mathbf{W}_{s1} \, \mathbf{H})) \textrm{,}
\end{equation}
where $r$ is the number of global attentions we want to extract from the sentence, and $\mathbf{W}_{s1}$ and $\mathbf{W}_{s2}$ are $s$-by-$d_{h}$ and  $r$-by-$s$ matrices, respectively. $s$ is a hyperparameter governing the dimensionality, and therefore maximum rank, of the attention-producing process. 

With the attention calculated, we compute an $r \times d_{h}$ embedding matrix, $\mathbf{M}=\mathbf{G}\mathbf{H}$, which in essence executes $r$ weighted sums across the $T$ hidden states, aggregating them together into $r$ representations. Each row of $\mathbf{G}$, denoted $\mathbf{g}^i$ $(i\in\{1\ldots r\})$, indicates how much each hidden state contributes to the final embedded representation of $\mathbf{M}$. We can thus draw a heat map for each row of the embedding matrix $M$ (See Figure \ref{fig:sampleResults} for examples). This way of visualization gives hints on what is encoded in each part of the embedding, adding an extra layer of interpretation.

To provide a final global text embedding of the sentences in the report, the AETE executes max-over-$r$ pooling across $\mathbf{M}$, producing an embedding vector $\hat{\mathbf{X}}_{AETE}$ with size $d_{h}$.

\subsection{Saliency Weighted Global Average Pooling}

In addition to using attention to provide a more meaningful text embedding, our goal is also to produce improved visual embeddings for classification. For this purpose, we re-use the attention mechanism, $\mathbf{G}$, except that we perform a max-over-$r$ operation, producing a sequence of saliency values, $g_t (t=1,\ldots,T)$, for each word, $\mathbf{w}_t$. These saliency values are used to weight and select the spatial attention maps, $\mathbf{a}_t$, generated at each time point:
\begin{equation}
\mathbf{a}_{ws}(x,y) = \sum_{t}{\mathbf{a}_{t}(x,y)*g_{t}} \textrm{.}
\end{equation}
This map is encoded with all spatial saliency regions guided by the text attention. We use this this map to highlight the spatial regions of $\mathbf{X}$ with more meaningful information:
\begin{equation}
\hat{\mathbf{X}}_{SW-GAP}(c) = \sum_{(x,y)}{\mathbf{a}_{ws}(x,y)*\mathbf{X}(x,y,c)} \textrm{,}
\end{equation}
where $x,y \in\{1...D\} $ and $\hat{\mathbf{X}}_{SW-GAP}$ is a 1-by-C vector representing the global visual information, guided by both text- and visual-based attention.  The lower part of figure \ref{fig:framework} illustrates an example of such pooling strategy. 

\subsection{Joint Learning}

With global representations computed for both the image and report, these must be combined together to produce the final classification. To accomplish this, we concatenate the two forms of representations $\hat{\mathbf{X}}=[\hat{\mathbf{X}}_{AETE};\hat{\mathbf{X}}_{SW-GAP}]$ and use a final fully-connected layer to produce the output for multi-label classification. The intuition behind our model is that the connection between the CNN and RNN network will benefit the training of both because the image activations can be adjusted for the text embedding task and salient image features could be extracted by pooling based on high text saliency.

In a similar fashion as Wang \etal{}~\cite{wang2017chestxa}, we define an $M$-dimensional disease label vector $\textbf{y} = [y_{1},...,y_{m},...,y_{M}], y_{m}\in\{0,1\}$ for each case and $M=15$ indicates the number of classes.
% * <adam.p.harrison@gmail.com> 2017-11-15T02:38:52.888Z:
% 
% > C=15
% Should this be 14?
% 
% ^.
$y_{m}$ indicates the presence with respect to a pathology or `no finding' (of listed disease categories) in the image. Here, we adopt the NLP-mined labels provided by~\cite{wang2017chestxa} as the `ground-truth' during the training.

The instance numbers for different disease categories are highly unbalanced, from hundreds to dozens of thousands.
In addition to the positive/negative balancing introduced in~\cite{wang2017chestxa}, we add weights to instances associated with different categories,
{\small
\begin{align}
L_{m} (f(I,\mathbf{S}),\mathbf{y}) & = \beta_{P}\sum_{y_{m}=1}-\ln(f(I,\mathbf{S}))\cdot \lambda_{m} \nonumber\\
& \quad + \beta_{N}\sum_{y_{m}=0}-\ln(1-f(I,\mathbf{S}))\cdot \lambda_{m} \textrm{,} %\label{eq:SCEL-weighted}
\label{eq:balanced}
\end{align}}
where $\beta_{P}=\frac{|N|}{|P|+|N|}$ and $\beta_{N}=\frac{|P|}{|P|+|N|}$.
$|P|$ and $|N|$ are the total number of images with at least one disease and with  no diseases, respectively.
$\lambda_{m} = (Q-Q_{m})/Q$  is a set of precomputed class-wised weights, where $Q$ and $Q_{m}$ are the total number of images and the number of images that have disease label $m$. $\lambda_{m}$ will be larger if the number of instances from class $m$ is small.

Because the TieNet can also generate text reports, we also optimize the RNN generative model loss~\cite{xu2015show}, $L_{R}$. Thus the overall loss is composed of two parts, the sigmoid cross entropy loss $L_{C}$ for the multi-label classification and the loss $L_{R}$ from the RNN generative model~\cite{xu2015show},
\begin{equation}
L_{overall}=\alpha L_{C} + (1-\alpha) L_{R}
\label{eq:loss_all}
\end{equation}
where $\alpha$ is added to balance the large difference between the two loss types. 

%\section{Applications}

\subsection{Medical Image Auto-Annotation}
%During training, we minimize the sum of losses induced by pairs of images $I_m$ with corresponding reports $S_m$.

%\subsection{Baseline CNN-RNN encoder-decoder model}
%\label{sec:baseline}

One straightforward application of the TieNet is the auto-annotation task to mine image classification labels.
By omitting the generation of sequential words, we accumulate and back-propagate only the classification loss for better text-image embeddings in image classification.
Here, we use the NLP-mined disease labels as `ground truth' in the training.
Indeed we want to learn a mapping between the input image-report pairs and the image labels.
The report texts often contain more easy-to-learn features than the image side.
The contribution of both sources to the final classification prediction should be balanced via either controlling the feature dimensions or drop-off partial of the `easy-to-learn' data during training.
%We also experiment how different amount of annotated data will affect the testing performance. It will help determine how much annotation work is required when radiologists are available for hand-labeling small amount of selective data.

\subsection{Automatic Classification and Reporting of Thorax Diseases}

For a more difficult but real-world scenario, we transform the image-text embedding network to serve as a unified system of image classification and report generation when only the unseen image is available.
During the training, both image and report are fed and two separate losses are computed as stated above, \ie{}, the loss for image classification and the loss for sequence-to-sequence modeling.
While testing, only the image is required as the input.
The generated text contained the learned text embedding recorded in the LSTM units and later used in the final image classification task.
The generative model we integrated into the text-image embedding network is the key to associate an image with its attention encoded text embedding.
%Additionally, the localization of disease in the images is implemented in a similar fashion to~\cite{wang2017chestxa,zhou2016learning}.
%The parameters in the final classification layer are splitted into two parts $\big[W_{text}, W_{image}\big], dim(W_{image})=C\times15$.
%Heatmaps for each disease categories could be generated by multiplying the activation $X$ from Transition layer with $W_{image}$.

%------------------------------------------------------------------------
%\section{Experiments}

%We define the experimental setup in Section~\ref{sec:exp setup}, and present the results in Section~\ref{sec:exp results}.

%\subsection{Experimental setup}
%\label{sec:exp setup}

\section{Dataset} 
\textbf{ChestX-ray14}~\cite{wang2017chestxa} is a recently released benchmark dataset for common thorax disease classification and localization.
It consists of 14 disease labels that can be observed in chest X-ray, \ie{}, Atelectasis, Cardiomegaly, Effusion, Infiltration, Mass, Nodule, Pneumonia, Pneumothorax, Consolidation, Edema, Emphysema, Fibrosis, Pleural Thickening, and Hernia.
%We obtain the entire dataset from National Institutes of Health Clinical Center through an inter-institutional agreement, including all the front-view chest X-ray images, anonymized radiological reports associated with each image and $\sim1000$ hand-labeled bounding boxes that indicates the location of disease patterns, in which images and bounding boxes are publicly available.
%We follow the NLP based label mining procedure introduced in~\cite{wang2017chestxa} and 
The NLP-mined labels are used as `ground truth' for model training throughout the experiments.
We adopt the patient-level data splits published with the data~\footnote{\url{https://nihcc.app.box.com/v/ChestXray-NIHCC}}.  

\textbf{Hand-labeled}: In addition to NLP-mined labels, we randomly select 900 reports from the testing set and have two radiologists to annotate the 14 categories of findings for the evaluation purpose.
A trial set of 30 reports was first used to synchronize the criterion of annotation between two annotators.
Then, each report was independently annotated by two annotators.
In this paper, we used the inter-rater agreement (IRA) to measure the consistency between two observers.
The resulting Cohen’s kappa is 84.3\%.
Afterwards, the final decision was adjudicated between two observers on the inconsistent cases.

\textbf{OpenI \cite{demner2015preparing}} is a publicly available radiography dataset collected from multiple institutes by Indiana University.
Using the OpenI API, we retrieved 3,851 unique radiology reports and 7,784 associated frontal/lateral images where each OpenI report was annotated with key concepts (MeSH words) including body parts, findings, and diagnoses.
For consistency, we use the same 14 categories of findings as above in the experiments.
In our experiments, only 3,643 unique front view images and corresponding reports are selected and evaluated.

\section{Experiments}
\textbf{Report vocabulary:}
We use all 15,472 unique words in the training set that appear at least twice.
Words that appear less frequently are replaced by a special out-of-vocabulary token, and the start and the end of the reports are marked with a special $\langle$START$\rangle$ and $\langle$END$\rangle$ token.
The pre-trained word embedding vectors was learned on PubMed articles using the gensim word2vec implementation with the dimensionality set to 200~\footnote{\url{https://radimrehurek.com/gensim/models/word2vec.html}}.
The word embedding vectors will be evolved along with other LSTM parameters.

\textbf{Evaluation Metrics:} To compare previous state-of-the-art works, we choose different evaluation metrics for different tasks so as to maintain consistency with data as reported in the previous works.

Receiver Operating Curves (ROC) are plotted for each disease category to measure the image classification performance and afterward, Areas Under Curve (AUC) are computed, which reflect the overall performance as a summary of different operating points.
%In addition, precision and recall are also provided for showing the actual testing performances if the framework is deployed.

To assess the quality of generated text report, BLEU scores~\cite{papineni2002bleu}, METEOR~\cite{banerjee2005meteor} and ROUGE-L~\cite{lin2004rouge} are computed between the original reports and the generated ones.
Those measures reflect the word overlapping statistics between two text corpora.
However, we believe their capabilities are limited for showing the actual accuracy of disease words (together with their attributes) overlapping between two text corpora.

\textbf{Training:}
The LSTM model contains a 256 dimensional cell and $s=2000$ in $\mathbf{W}_{s1}$ and $\mathbf{W}_{s2}$ for generating the attention weights $\mathbf{G}$.
During training, we use 0.5 dropout on the MLP and 0.0001 for L2 regularization.
We use the Adam optimizer with a mini-batch size of 32 and a constant learning rate of 0.001.

In addition, our self-attention LSTM has a hidden layer with
350 units. %(the $x$ in Section~\ref{sec:sentence emb})
We choose the matrix embedding to have 5 rows (the $r$), and a coefficient of 1 for the penalization term.
All the models are trained until convergence is achieved and the hyper-parameters for testing is selected according to the corresponding best validation set performance.

Our text-image embedding network is implemented based on TensorFlow~\cite{abadi2016tensorflow} and Tensorpack~\footnote{\url{https://github.com/ppwwyyxx/tensorpack/}}.
The ImageNet pre-trained model, \ie{}, ResNet-50~\cite{he2016deep} is obtained from the Caffe model zoo and converted into the TensorFlow compatible format.
The proposed network takes the weights from the pre-trained model and fixes them during the training.
Other layers in the network are trained from scratch.
In a similar fashion as introduced in~\cite{wang2017chestxa}, we reduce the size of mini-batch to fit the entire model in each GPU while we accumulate the gradients for a number of iterations and also across a number of GPUs for better training performance.
The DCNN models are trained using a Dev-Box Linux server with 4 Titan X GPUs.

{\small \small
\begin{table*}[t]
	\begin{center}
		\begin{tabularx}{\textwidth}{Xr@{~/~}r@{~/~}r@{~/~}r@{~/~}rr@{~/~}r@{~/~}r@{~/~}r@{~/~}rr@{~/~}r@{~/~}r@{~/~}r@{~/~}r}
			\toprule
			\multirow{2}{*}{Disease} & \multicolumn{5}{c}{ChestX-ray14} & \multicolumn{5}{c}{Hand-labeled}&\multicolumn{5}{c}{OpenI}\\
			\cmidrule(rl){2-6}\cmidrule(rl){7-11}\cmidrule{12-16}
			&R & I+R & I \cite{wang2017chestxa}& I+GR & \# &R & I+R & I \cite{wang2017chestxa}& I+GR& \# &R & I+R & I \cite{wang2017chestxa}& I+GR & \#\\\midrule
			%\multicolumn{7}{c}{OpenI}\\
			Atelectasis 	&.983	&.993	&.700	&.732	&3255	&.886	&.919	&.680	&.715	&261	&.981	&.976	&.702	&.774	&293\\
			Cardiomegaly 	&.978	&.994	&.810	&.844	&1065	&.964	&.989	&.820	&.872	&185	&.944	&.962	&.803	&.847	&315\\
			Effusion 		&.984	&.995	&.759	&.793	&4648	&.938	&.967	&.780	&.823	&257	&.968	&.977	&.890	&.899	&140\\
			Infiltration 	&.960	&.986   &.661	&.666	&6088	&.849	&.879	&.648	&.664	&271	&.981	&.984	&.585	&.718	&57\\
			Mass 			&.984	&.994	&.693	&.725	&1712	&.935	&.943	&.696	&.710	&93		&.959	&.903	&.756	&.723	&14\\
			Nodule 			&.981	&.994	&.668	&.685	&1615	&.974	&.974	&.662	&.684	&130	&.967	&.960	&.647	&.658	&102\\
			Pneumonia 		&.947	&.969	&.658	&.720	&477	&.917	&.946	&.724	&.681	&55		&.983	&.994	&.642	&.731	&36\\
			Pneumothorax 	&.983	&.995	&.799	&.847	&2661	&.983	&.996	&.784	&.855	&166	&.960	&.960	&.631	&.709	&22\\
			Consolidation 	&.989	&.997	&.703	&.701	&1815	&.923	&.910	&.609	&.631	&60		&.969	&.989	&.790	&.855	&28\\
			Edema 			&.976	&.989	&.805	&.829	&925	&.970	&.987	&.815	&.834	&33		&.984	&.995	&.799	&.879	&40\\
			Emphysema 		&.996	&.997	&.833	&.865	&1093	&.980	&.981	&.835	&.863	&44		&.849	&.868	&.675	&.792	&94\\
			Fibrosis 		&.986	&.986	&.786	&.796	&435	&.930	&.989	&.688	&.714	&11		&.985	&.960	&.744	&.791	&18\\
			PT 				&.988	&.997	&.684	&.735	&1143	&.904	&.923	&.679	&.776	&41		&.948	&.953	&.691	&.749	&52\\
			Hernia 			&.929	&.958	&.871	&.876	&86 	&.757	&.545	&.864	&.647	&2		&-- 	&-- 	&-- 	&--		&0\\
			NoFinding 		&.920	&.985	& -- 	&.701	&9912	&.889	&.908	&--		&.666	&85		&.933	&.936	&--		&.747	&2789\\
			\midrule
			\textit{AVG} 	&.976	&.989	&.745	&.772	&--	&.922	&.925	&.735	&.748	&--	&.960	&.965	&.719	&.779	&--\\
			\midrule
			\textit{\#wAVG} &.978	&.992	&\textbf{.722}	&\textbf{.748}	&--	&.878	&.900	&\textbf{.687}	&\textbf{.719}	&--	&.957	&.966	&\textbf{.741}	&\textbf{.798}	&--	\\
			\bottomrule
		\end{tabularx}
	\end{center}
	\caption{Evaluation of image classification results (AUCs) on ChestX-ray14, hand-labeled and OpenI dataset. Performances are reported on four methods, \ie{}, multilabel classification based on Report (R), Image + Report (I+R), Image ~\cite{wang2017chestxa}, and Image + Generative Report(I+GR).}
	\label{tab:classification_AUC}
\end{table*}
}
\begin{figure*}[t]
	\centering
	\begin{tabular}{cc}
		\includegraphics[width=0.32\linewidth]{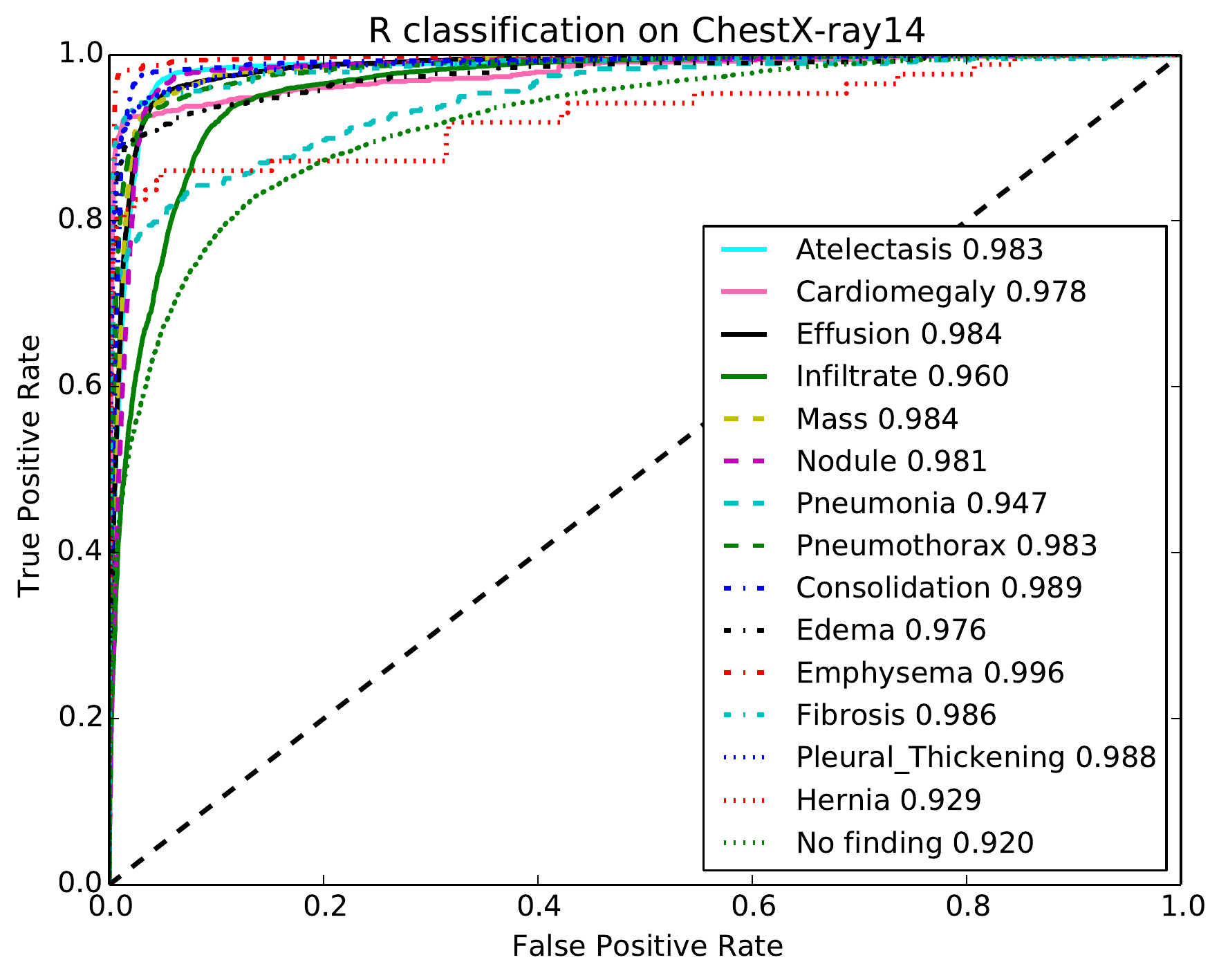} 
		\includegraphics[width=0.32\linewidth]{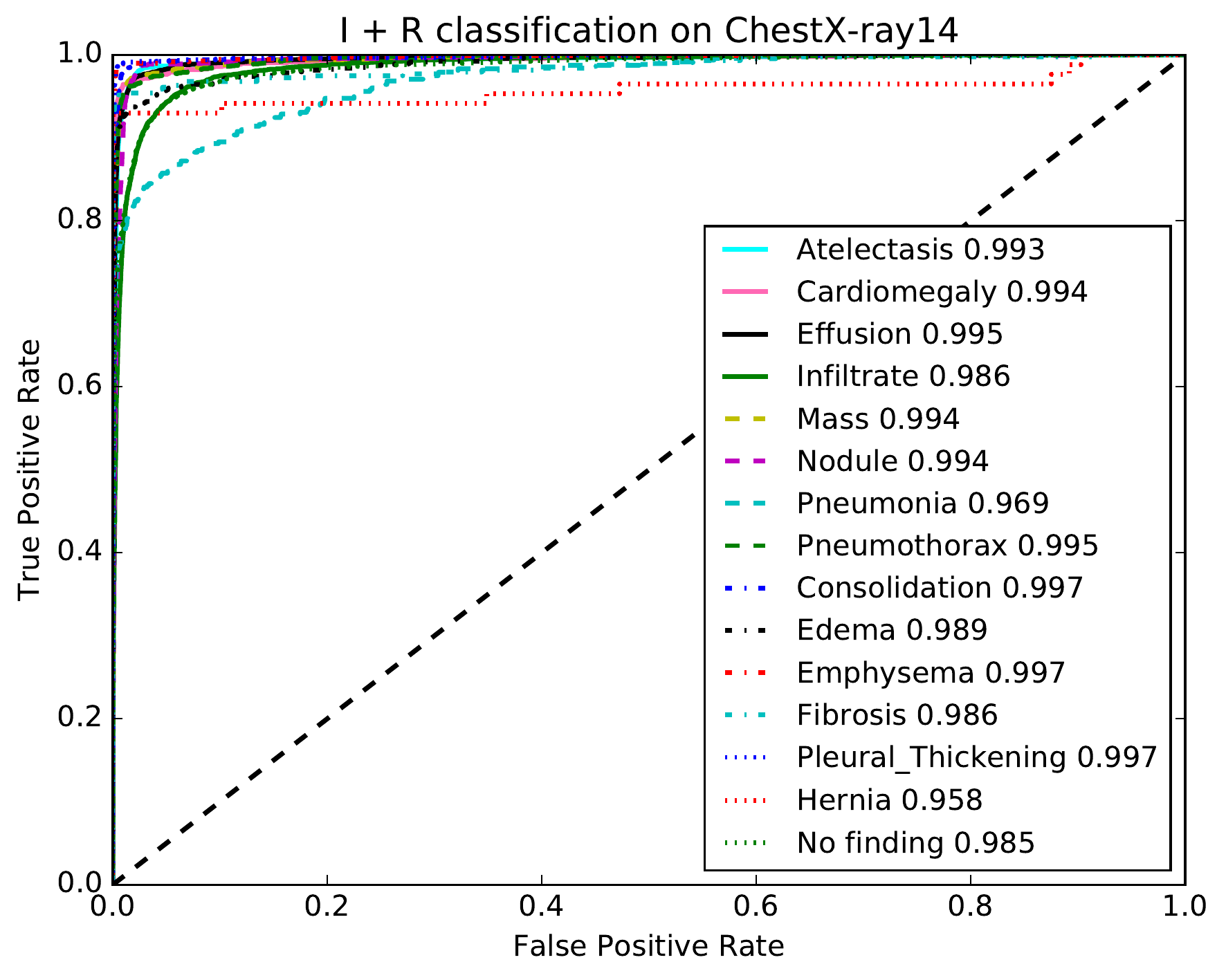} 
		\includegraphics[width=0.32\linewidth]{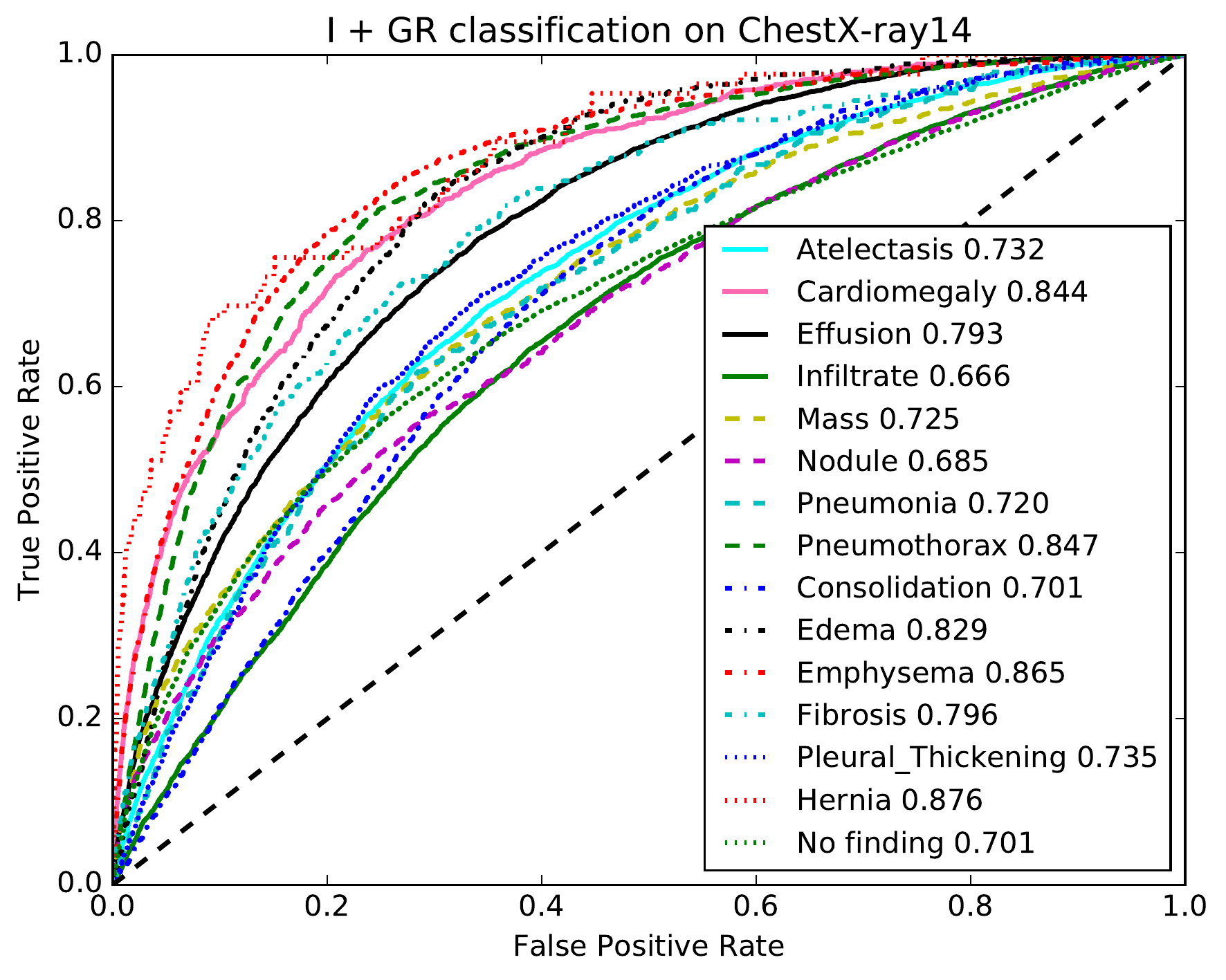} \\
		\includegraphics[width=0.32\linewidth]{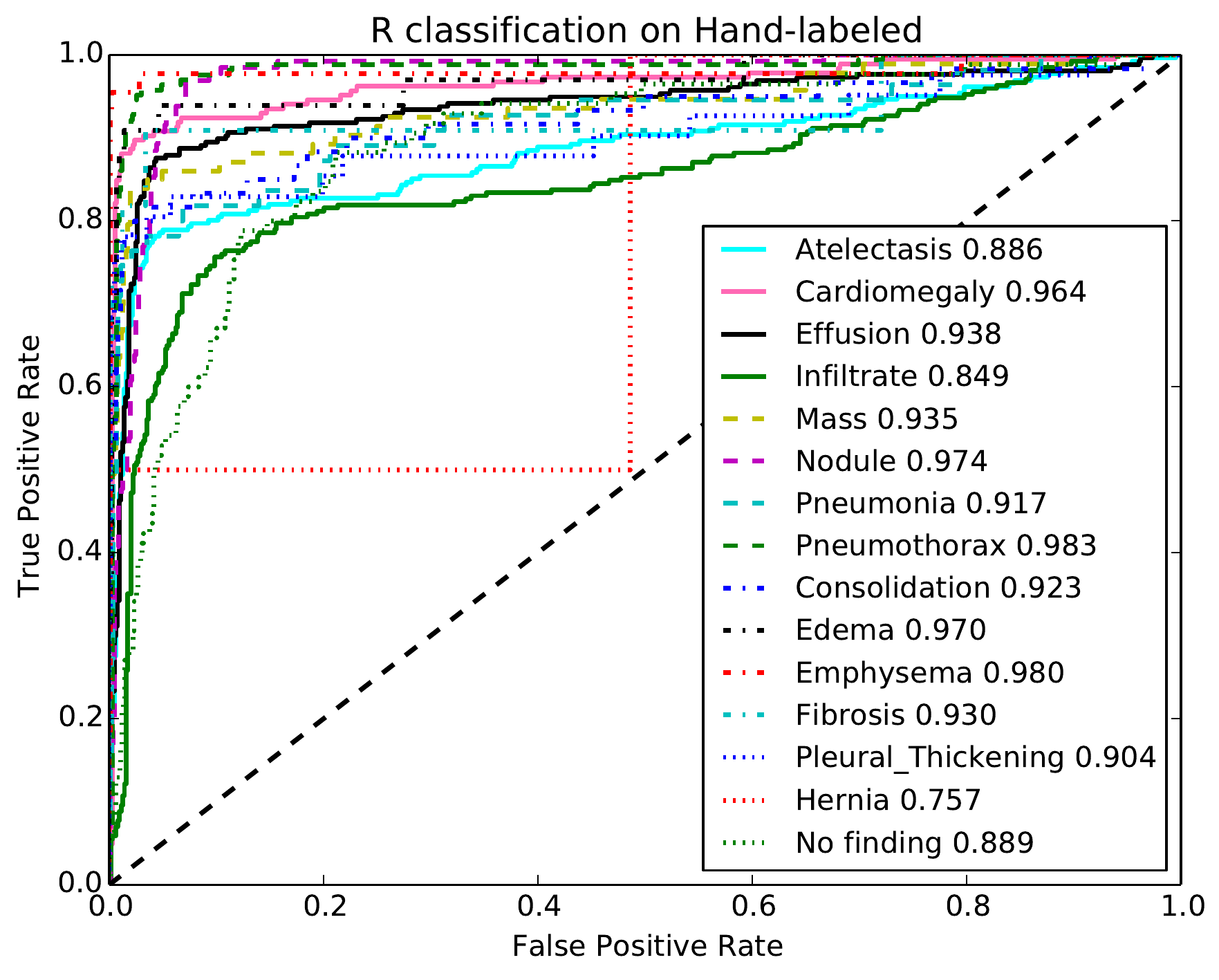} 
		\includegraphics[width=0.32\linewidth]{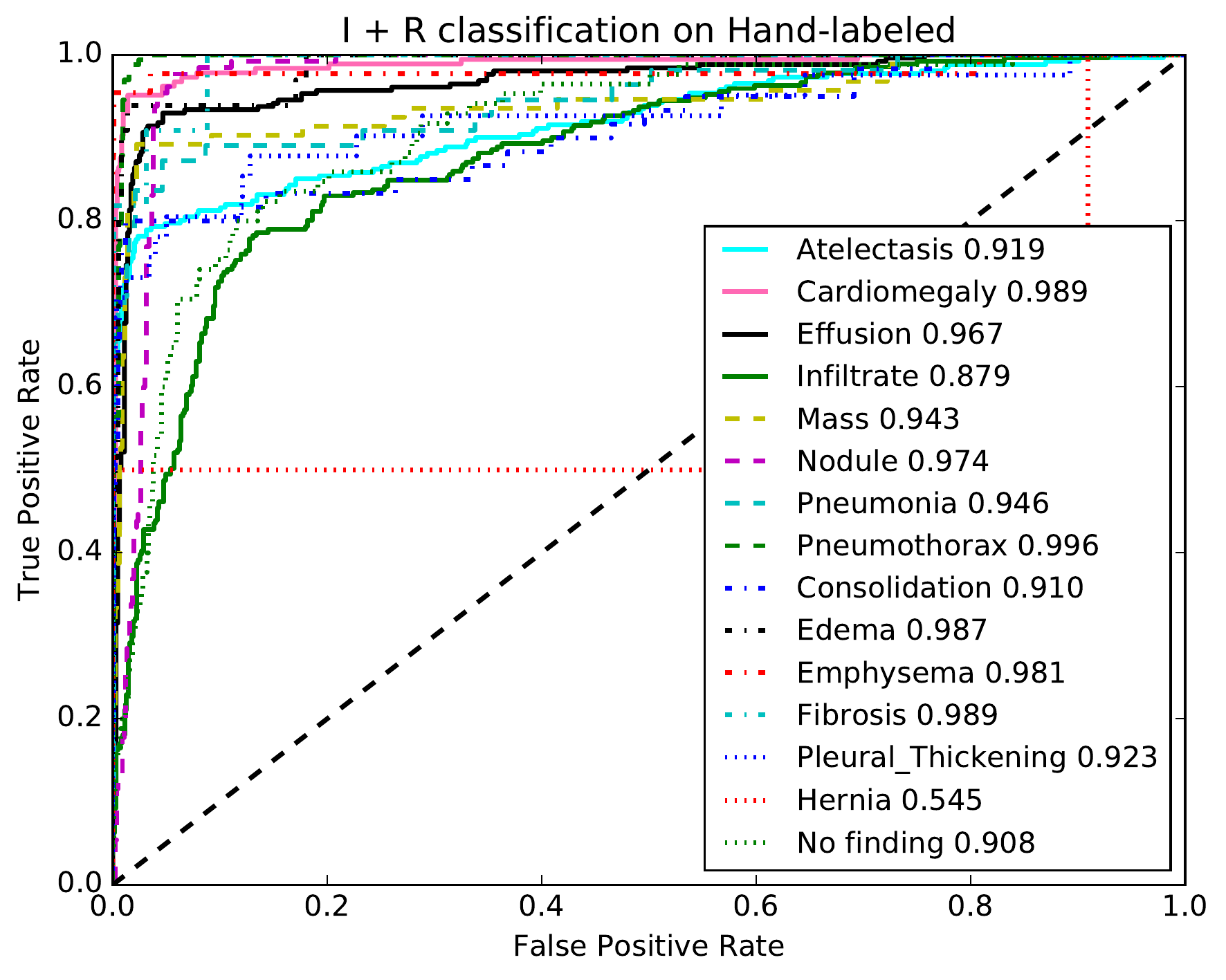} 
		\includegraphics[width=0.32\linewidth]{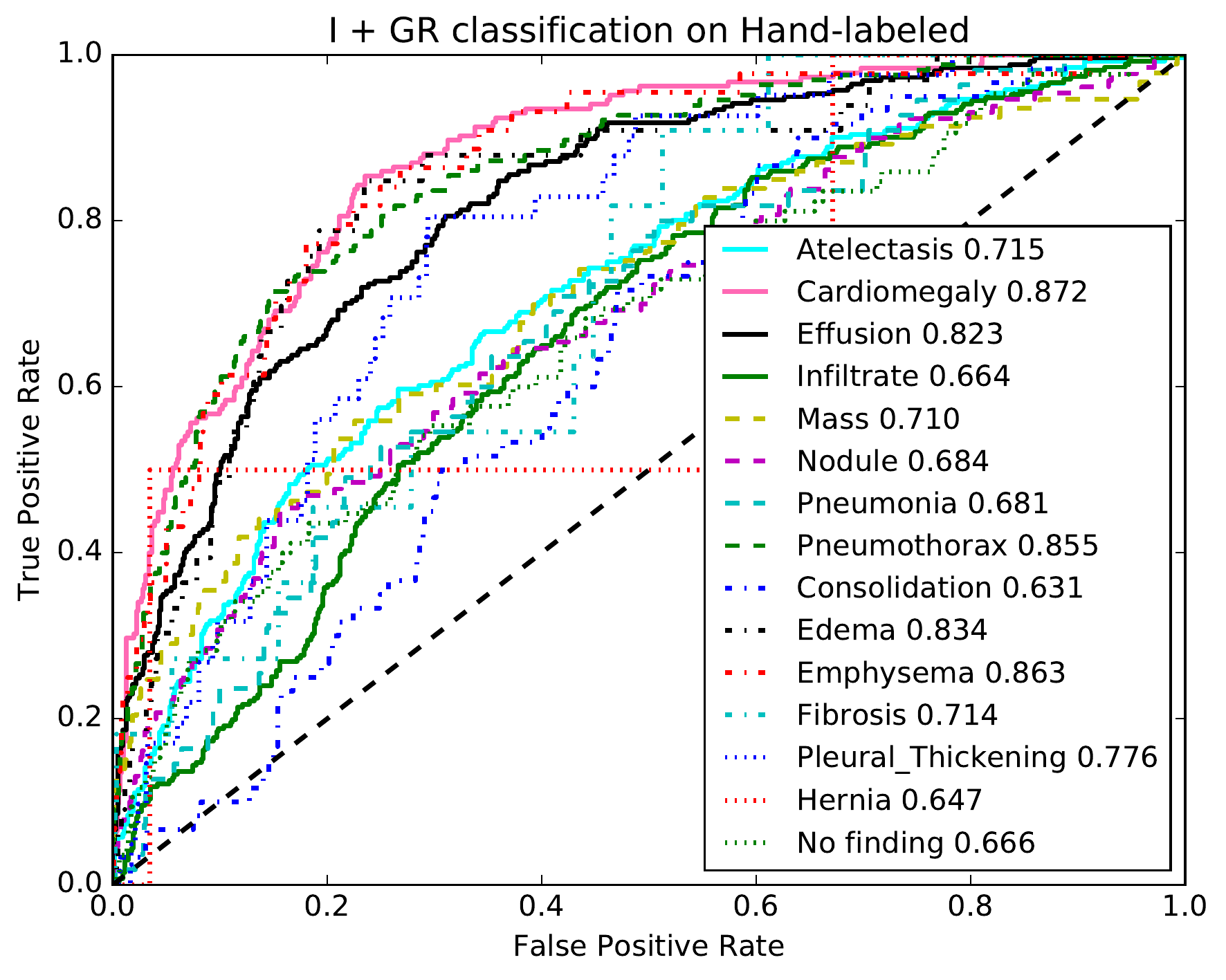} \\
	\end{tabular}
	\caption{A comparison of classification performance with different testing inputs, \ie{} Report (R), Image+Report (I+R), and Image+Generative Report(I+GR).}
	\label{fig:ROC_3set}
	\vspace*{-1em}
\end{figure*}

\begin{table}[t]
	%\vspace*{-1em}
	\caption{Evaluation of generated reports in ChestX-ray14 testing set using BLEU, METEOR and ROUGE-L.}
	\centering
	\begin{tabular}{lrr}%{\columnwidth}{Xrr}
		\toprule
		%\multicolumn{3}{|c|}{ChestX-ray14 Testing Set}\\\hline \hline
		{\bf }   & {Captioning~\cite{xu2015show}} & {TieNet I+GR} \\%& {\bf Output}\\[0.1ex]
		\midrule
		{ BLEU-1}   & 0.2391& 0.2860\\ 
		{ BLEU-2}    & 0.1248& 0.1597\\
		{ BLEU-3}   & 0.0861& 0.1038 \\
		{ BLEU-4}   & 0.0658& 0.0736 \\
		{ METEOR}    & 0.1024& 0.1076\\
		{ ROUGE-L}    & 0.1988&0.2263\\	
		\bottomrule
	\end{tabular}\label{tab:eval_report}
%	\vspace*{-1em}
\end{table}

\subsection{Auto-annotation of Images}
\label{sec:exp results}
Figure \ref{fig:ROC_3set} illustrates the ROC curves for the image classification performance with 3 different inputs evaluated on 3 different testing sets, \ie{}, ChestX-ray14 testing set (ChestX-ray14), the hand-labeled set (Hand-labeled) and the OpenI set (OpenI).
Separate curves are plotted for each disease categories and `No finding'.
Here, two different auto-annotation frameworks are trained by using different inputs, \ie{}, taking reports only (R) and taking image-report pairs (I+R) as inputs.
When only the reports are used, the framework will not have the saliency weighted global average pooling path.
In such way, we can get a sense how the features from text path and image path individually contribute to the final classification prediction.

We train the proposed auto-annotation framework using the training and validation sets from the ChestX-ray14 dataset and test it on all three testing sets, \ie{}, ChestX-ray14, hand-labeled and OpenI.
Table \ref{tab:classification_AUC} shows the AUC values for each class computed from the ROC curves shown in Figure \ref{fig:ROC_3set}.
The auto-annotation framework achieves high performance on both ChestX-ray14 and Hand-labeled, \ie{}, over 0.87 in AUC with reports alone as the input and over 0.90 in AUC with image-report pairs on sample number weighted average ($\#wAVG$).
The combination of image and report demonstrates the supreme advantage in this task.
In addition, the auto-annotation framework trained on ChestX-ray14 performed equivalently on OpenI. It indicates that the model trained on a large-scale image dataset could easily be generalized to the unseen data from other institutes.
%It may be due to the large difference of reporting style amongst radiologists and institutes.
%In this case, our TieNet learned on the reports from one particular institute may not be applied successfully to the data from others.
The model trained solely based on images could also be generalized well to the datasets from other sources.
In this case, both the proposed method and the one in ~\cite{wang2017chestxa} are able to perform equally well on all three testing sets.

\begin{figure*}[t]
	\centering
	\includegraphics[width=1\linewidth,clip,trim=.4cm 3.3cm 4.6cm 0.4cm]{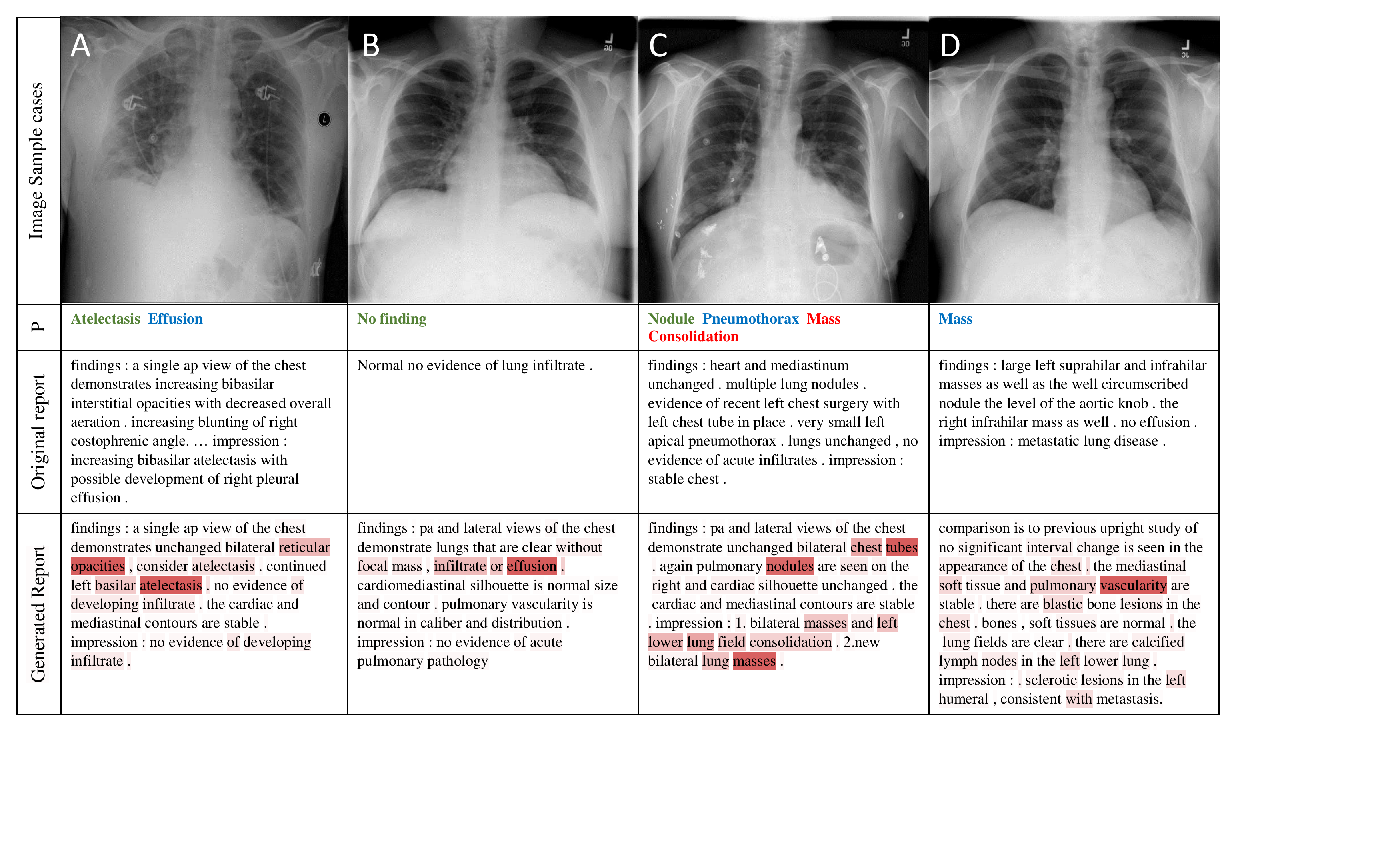}
	\caption{4 sample image Classification Predictions (P) along with original and generated reports. Text attentions are highlighted over the generated text. Correct predication is marked in green, false prediction in red and missing prediction in blue.}
	\label{fig:sampleResults}
\end{figure*}

\subsection {Classification and Reporting of Chest X-ray}
When the TieNet is switched to an automatic disease classification and reporting system, it takes a single image as the input and is capable of outputting a multi-label prediction and corresponding radiological report together.
The ROC curves on the right in Figure \ref*{fig:ROC_3set} and Table 1 show the image classification performance produced by the multi-purpose reporting system.
The AUCs from our TieNet (I+GR) demonstrate the consistent improvement in AUCs ($2.3\%-5.7\%$ on $\#wAVG$ for all the disease categories) across all three datasets.
The multilabel classification framework~\cite{wang2017chestxa} serves as a baseline model that also takes solely the images.
Furthermore, the performance improvement achieved on the Hand-labeled and OpenI datasets (with ground truth image labels) is even larger than the performance gain on ChestX-ray14 (with NLP-mined labels).
It indicates that the TieNet is able to learn more meaningful and richer text embeddings directly from the raw reports and correct the inconsistency between embedded features and erroneous mined labels.

Table \ref{tab:eval_report} shows that the generated reports from our proposed system obtain higher scores in all evaluation metrics in comparison to the baseline image captioning model \cite{xu2015show}.
%The main difference between those two models is the image feature maps that are used to initialize the LSTM and inputted to the LSTM unit along with the word embedding at each time point.
It may be because the gradients from RNN are backpropagated to the CNN part and the adjustment of image features from Transition layer will benefit the report generation task.

Figure \ref{fig:sampleResults} illustrates 4 sample results from the proposed automatic classification and reporting system. Please see more examples in the appendix \ref{sec:app_A}.
Original images are shown along with the classification predications, original reports and generated reports. Text-attended words are also highlighted over the generated reports. If looking at generated reports alone, we find that they all read well. However, the described diseases may not truly appear in the images. For example, `Atelectasis' is correctly recognized in sample A but `Effusion' is missed. `Effusion' (not too far from the negation word `without') is erroneously highlighted in sample B but the system is still able to correctly classify the image as `No finding'. In sample D, the generated report misses `Mass' while it states right about the metastasis in the lung. One promising finding is that the false predictions (`Mass' and `Consolidation') in sample C can actually be observed in the image (verified by a radiologist) but somehow did not noted in the original report, which indicates our proposed netowrk can in some extent associate the image appearance with the text description. 

%\subsection{Analysis and Discussion}
%\textbf{image vs text}:

%------------------------------------------------------------------------
\section{Conclusion}
Automatically extracting the machine-learnable annotation from the retrospective data remains a challenging task, among which images and reports are two main useful sources.
Here, we proposed a novel text-image embedding network integrated with multi-level attention models.
TieNet is implemented in an end-to-end CNN-RNN architecture for learning a blend of distinctive image and text representations.
Then, we demonstrate and discuss the pros and cons of including radiological reports in both auto-annotation and reporting tasks.
While significant improvements have been achieved in multi-label disease classification, there is still much space to improve the quality of generated reports.
For future work, we will extend TieNet to include multiple RNNs for learning not only disease words but also their attributes and further correlate them and image findings with the description in the generated reports.

{\bf Acknowledgements } This work was supported by the Intramural Research Programs of the NIH Clinical Center and National Library of Medicine. Thanks to Adam Harrison and Shazia Dharssi for proofreading the manuscript. We are also grateful to NVIDIA Corporation for the GPU donation.

{\small
	\bibliographystyle{ieee}
	\bibliography{egbib}
}

\appendix

\section{More Experiment Results} \label{sec:app_A}
In this section, we present 20 more classification and reporting results (case E-X) from the proposed TieNet in addition to the four examples (case A-D) shown in the main paper. Sample images are illustrated along with associated classification Predictions (P), original and generated reports. Text attentions are highlighted with different saturation levels over the generated text. Darker red means higher weights of the text attention. Correct classification predications are marked in green, false predictions in red and missed predictions in blue.

\begin{figure*}[p]
	\centering
	\includegraphics[height=0.9\linewidth,trim=3.1cm 1.3cm 2.6cm 0.4cm,angle =270]{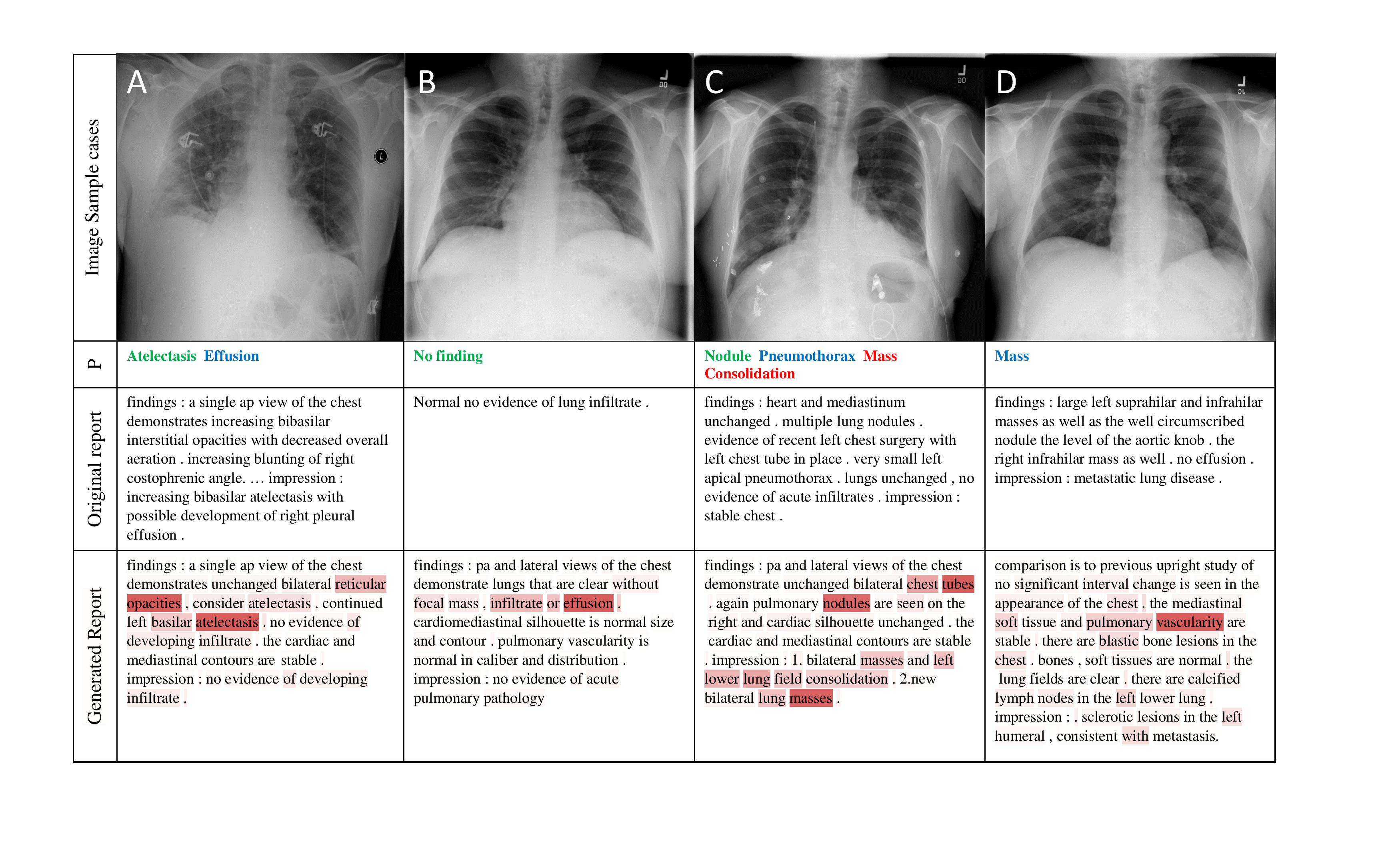}
	\caption{4 sample image Classification Predictions (P) along with original and generated reports. Text attentions are highlighted over the generated text. Correct predication is marked in green, false prediction in red and missing prediction in blue.}
	\label{fig:sampleResults}
\end{figure*}

\begin{figure*}[p]
	\centering
	\includegraphics[height=0.9\linewidth,trim=3.1cm 1.3cm 2.6cm 0.4cm,angle =270]{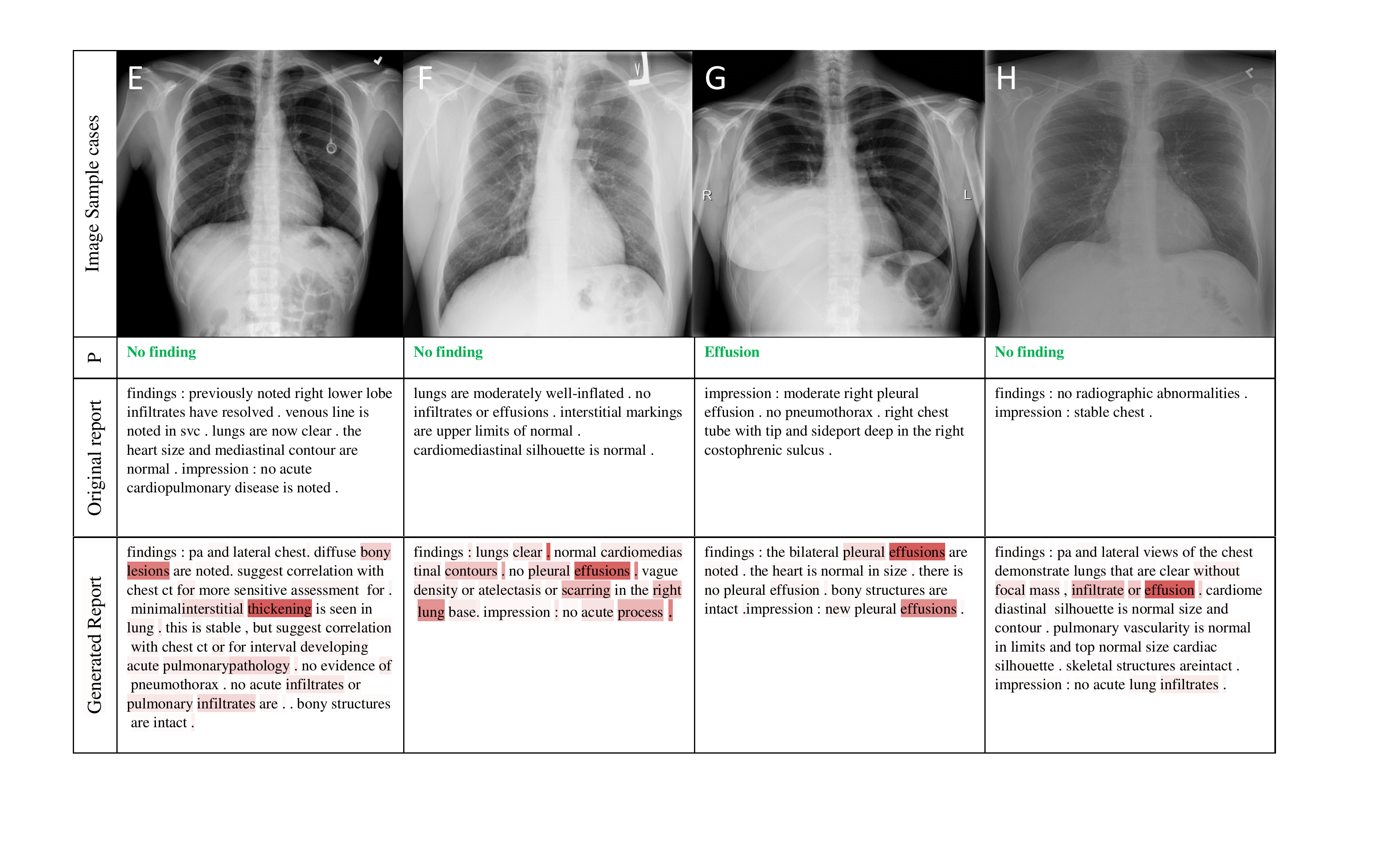}
	\caption{4 sample image Classification Predictions (P) along with original and generated reports. Text attentions are highlighted over the generated text. Correct predication is marked in green, false prediction in red and missing prediction in blue.}
	\label{fig:sampleResults}
\end{figure*}
\begin{figure*}[p]
	\centering
	\includegraphics[height=0.9\linewidth,trim=3.1cm 1.3cm 2.6cm 0.4cm,angle =270]{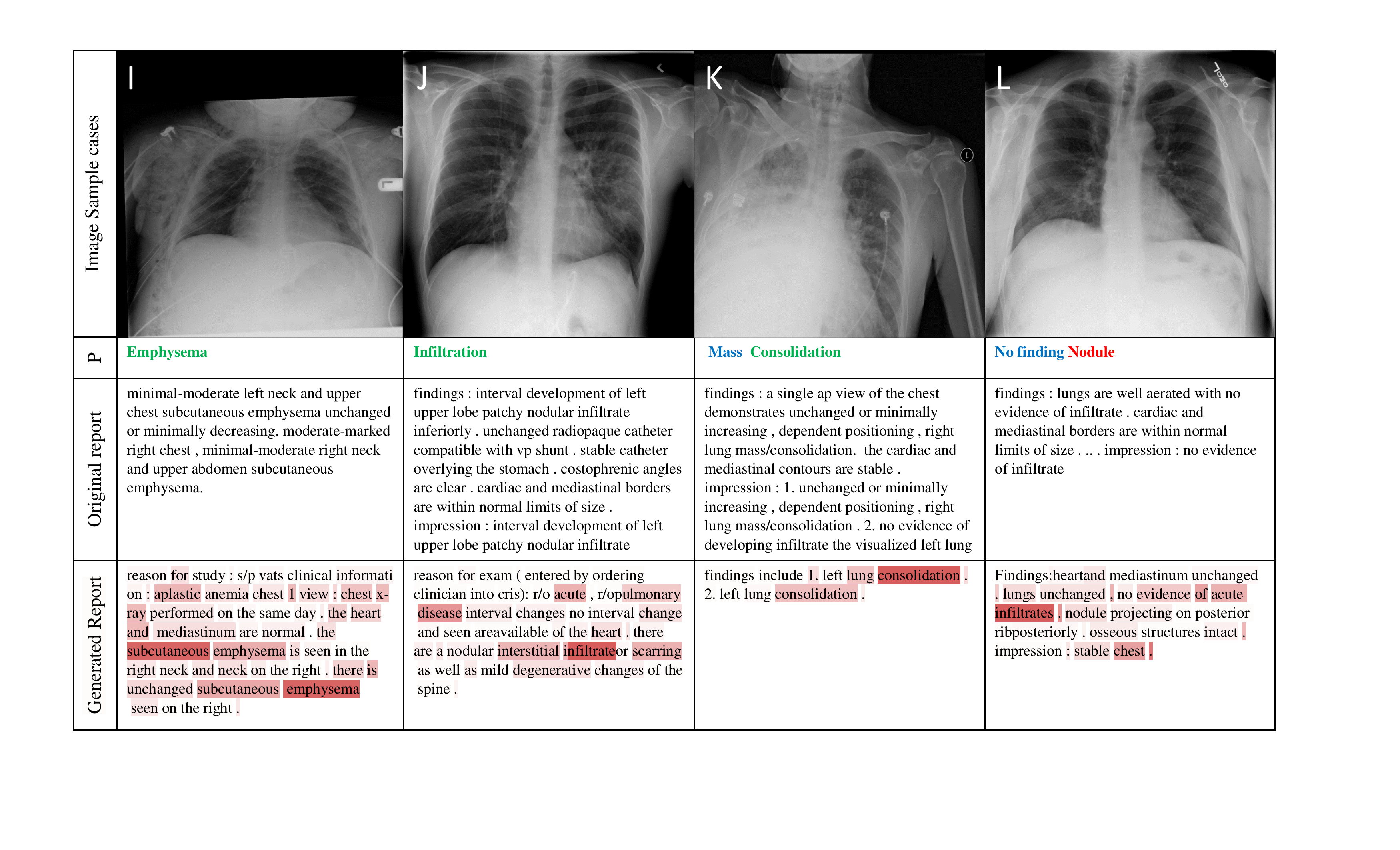}
	\caption{4 sample image Classification Predictions (P) along with original and generated reports. Text attentions are highlighted over the generated text. Correct predication is marked in green, false prediction in red and missing prediction in blue.}
	\label{fig:sampleResults}
\end{figure*}
\begin{figure*}[p]
	\centering
	\includegraphics[height=0.9\linewidth,trim=3.1cm 1.3cm 2.6cm 0.4cm,angle =270]{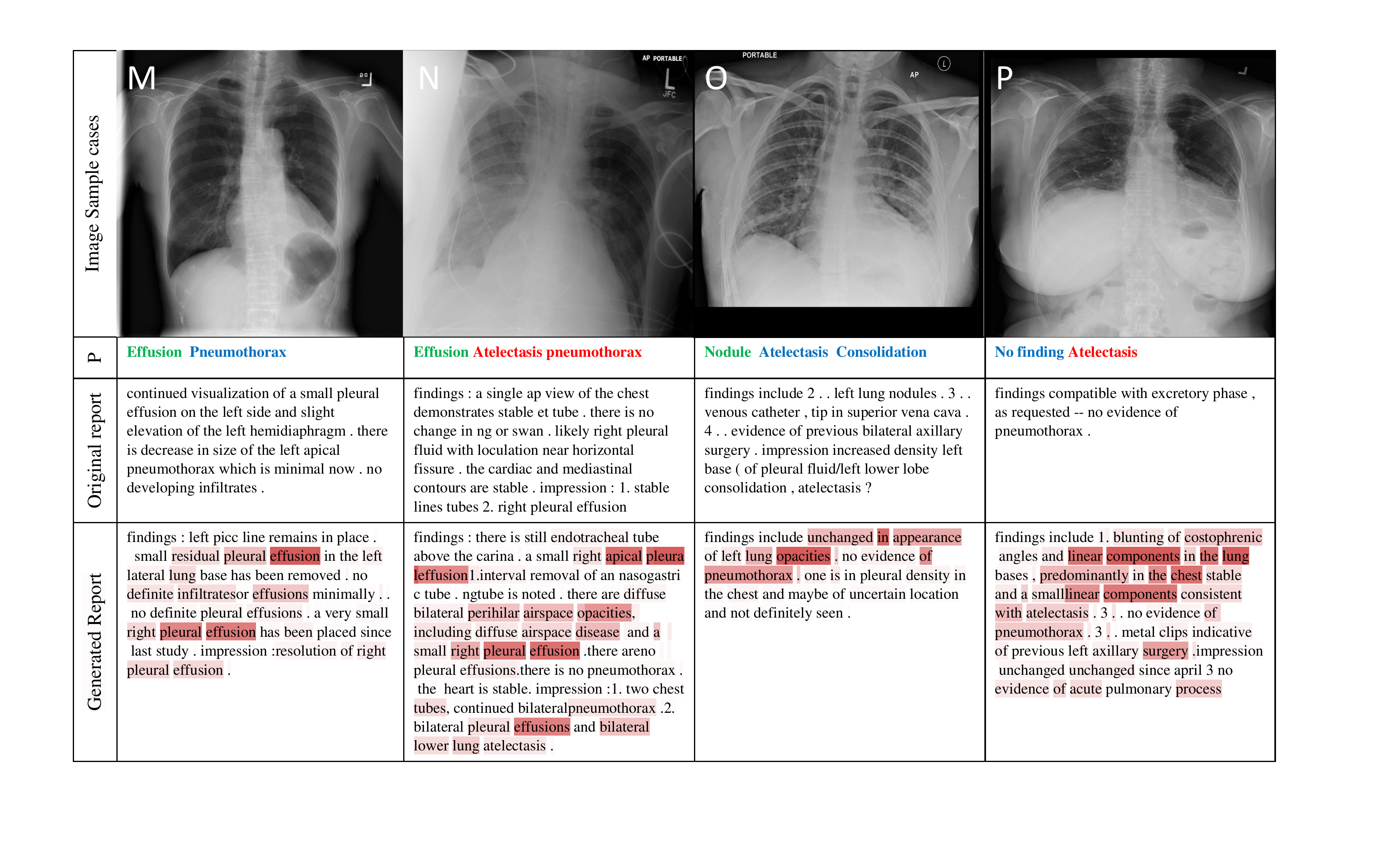}
	\caption{4 sample image Classification Predictions (P) along with original and generated reports. Text attentions are highlighted over the generated text. Correct predication is marked in green, false prediction in red and missing prediction in blue.}
	\label{fig:sampleResults}
\end{figure*}
\begin{figure*}[p]
	\centering
	\includegraphics[height=0.9\linewidth,trim=3.1cm 1.3cm 2.6cm 0.4cm,angle =270]{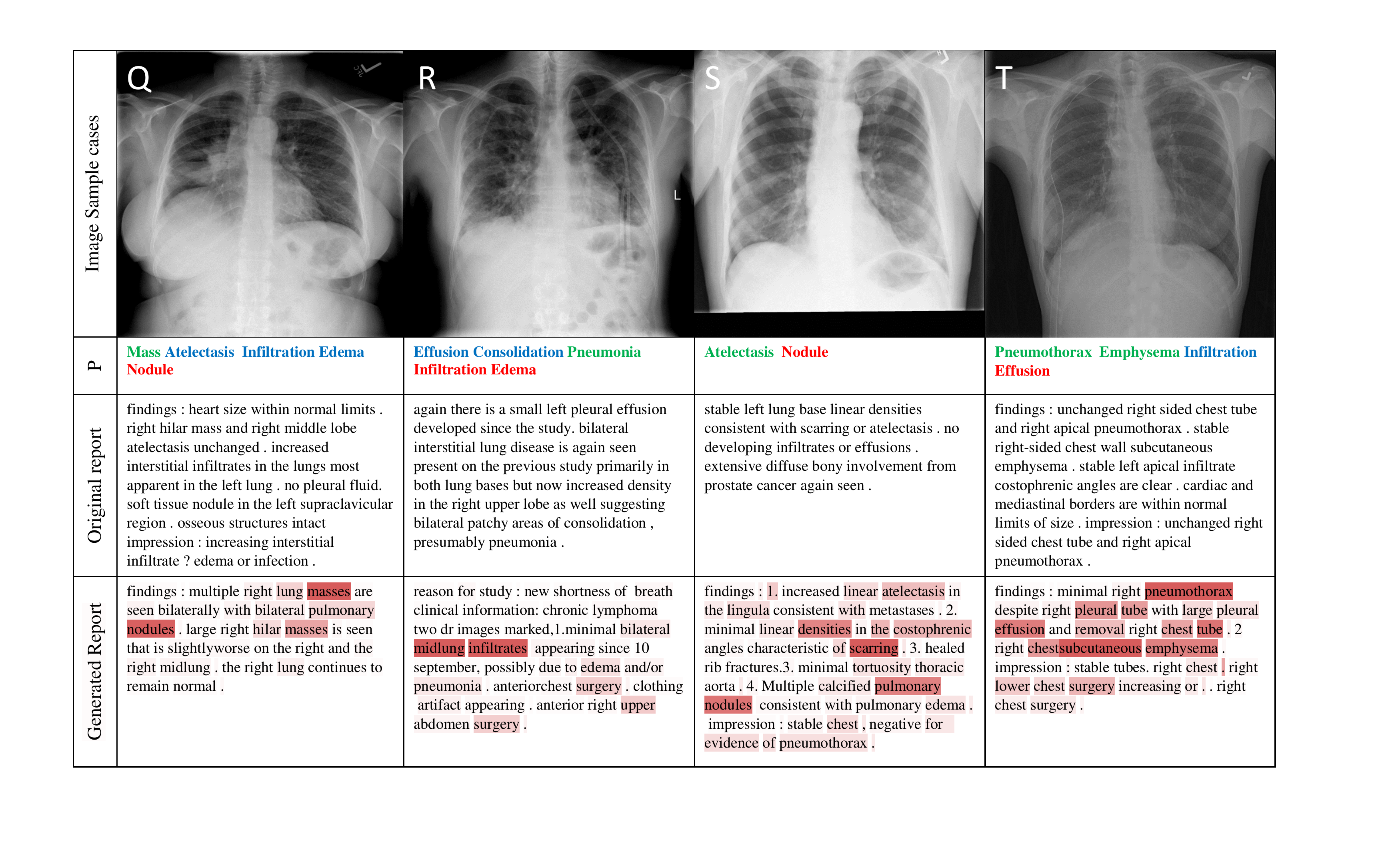}
	\caption{4 sample image Classification Predictions (P) along with original and generated reports. Text attentions are highlighted over the generated text. Correct predication is marked in green, false prediction in red and missing prediction in blue.}
	\label{fig:sampleResults}
\end{figure*}
\begin{figure*}[p]
	\centering
	\includegraphics[height=0.9\linewidth,trim=3.1cm 1.3cm 2.6cm 0.4cm,angle =270]{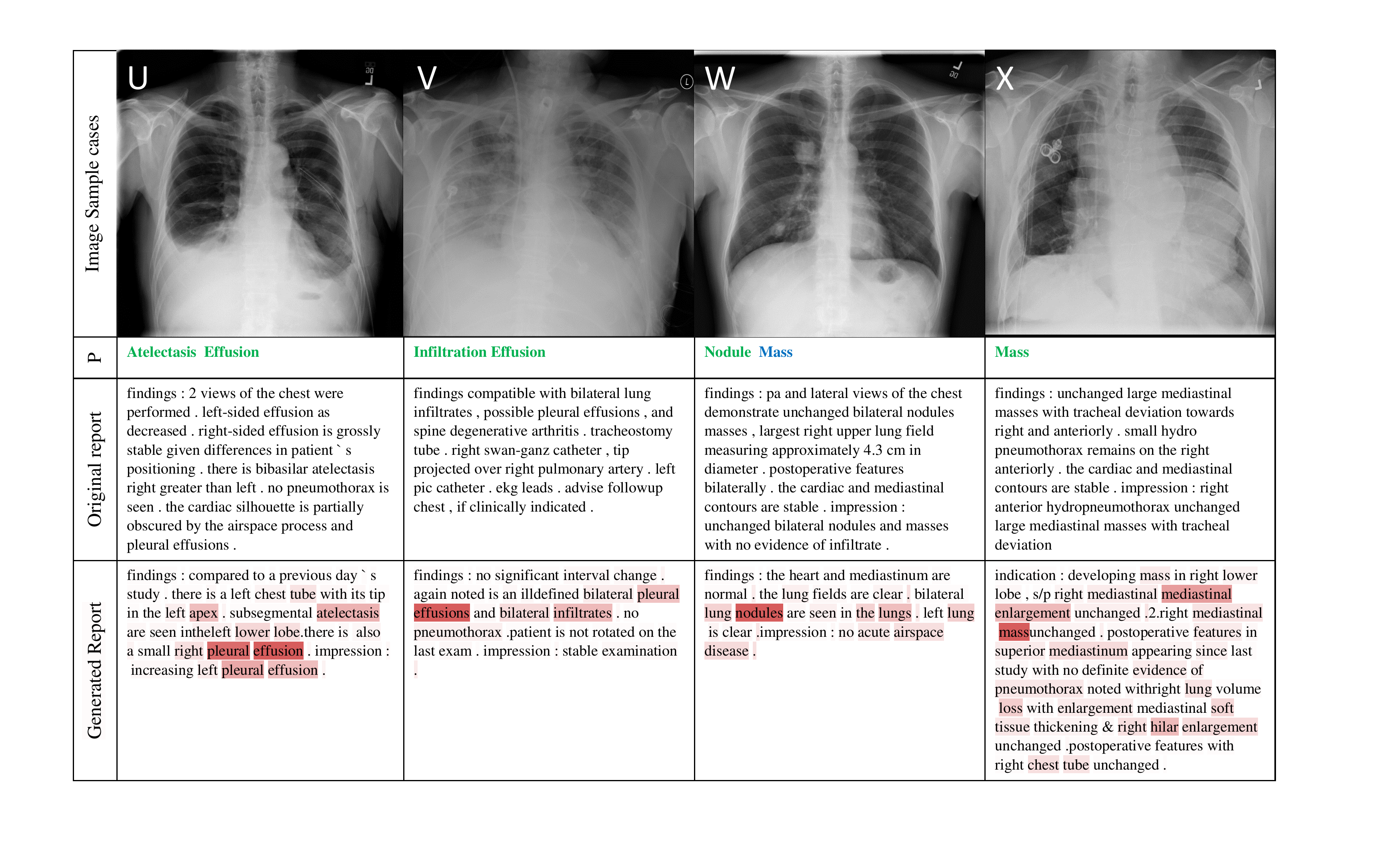}
	\caption{4 sample image Classification Predictions (P) along with original and generated reports. Text attentions are highlighted over the generated text. Correct predication is marked in green, false prediction in red and missing prediction in blue.}
	\label{fig:sampleResults}
\end{figure*}

\end{document}